\documentclass{article}
\PassOptionsToPackage{numbers,sort&compress}{natbib}
\usepackage[main,final]{neurips_2026}
\usepackage[utf8]{inputenc}
\usepackage[T1]{fontenc}
\usepackage{hyperref}
\usepackage{url}
\usepackage{graphicx}
\usepackage{microtype}
\usepackage{xcolor}
\usepackage{amsmath,amssymb,amsfonts}
\usepackage{booktabs,multirow,makecell,colortbl}
\usepackage{enumitem}
\usepackage{algorithm}
\usepackage{algpseudocode}
\usepackage{caption}
\usepackage{needspace}
\definecolor{polyured}{HTML}{8F1329}
\newcommand{\speedup}[1]{{\textcolor{polyured}{\bfseries #1}}}
\newcommand{\submetric}[1]{\,{\scriptsize\speedup{(#1)}}}
\newcommand{\codeurl}{\url{https://github.com/PolyU-VCLab/BitMTP}}
\hypersetup{hidelinks,
  pdftitle={BitMTP: When Multi-Token Prediction Meets Low-Bit Large Language Models},
  pdfauthor={Ning Zhang, Shihao Wang, Jinrui Zhang, Chaodong Xiao, Lei Zhang},
  pdfkeywords={multi-token prediction, ultra-low-bit LLM, BitNet, speculative decoding, CUDA Graph}}
\title{BitMTP: When Multi-Token Prediction Meets\\Low-Bit Large Language Models}
\author{Ning Zhang\quad Shihao Wang\quad Jinrui Zhang\quad Chaodong Xiao\quad
Lei Zhang\thanks{Corresponding author. This work is supported by the PolyU-OPPO Joint Innovative Research Center.}\\
Department of Computing, The Hong Kong Polytechnic University\\
\texttt{ning24.zhang@polyu.edu.hk,\quad cslzhang@comp.polyu.edu.hk}\\
\texttt{\{shihaow.wang,jin-rui.zhang,chaodong.xiao\}@connect.polyu.hk}}
\begin{document}
\maketitle
\begin{abstract}
For efficient deployment of large language models (LLMs), methods of ultra-low-bit backbones reduce weight storage, memory traffic, and linear-layer execution cost, while multi-token prediction (MTP) methods reduce sequential decoding steps by predicting multiple future tokens in parallel. Although complementary, the direct combination of these methods cannot accelerate decoding at comparable quality: masked drafting lacks intermediate token input supplied in next-token training, while wider computation and causal key--value (KV) updates can offset the savings from fewer steps. To address these challenges, we propose \textbf{BitMTP}, an algorithm--system co-design framework for efficient MTP on ultra-low-bit LLM backbones. A joint next-token prediction and MTP objective equips a single model with both autoregressive (AR) generation and block-causal drafting. Confidence-only decoding, speculative-only decoding, and hybrid decoding provide different quality--throughput operating points without using separate draft and verifier models. Accepted-prefix alignment keeps the committed prefix, materialized causal KV boundary, and CUDA Graph replay state consistent for continued generation. BitMTP uses this contract to select width-specialized low-bit execution and reuse captured graphs over shared KV storage. On NVIDIA H20, BitMTP with hybrid decoding achieves five-task mean speedups of $2.2\times$ on BitNet~b1.58-2B and $2.1$--$2.4\times$ on 1.7B--4B Bonsai models at comparable task quality. On RTX~3090 and Jetson Orin, it further achieves up to $2.0\times$ in five-task mean throughput and $3.3\times$ on MBPP+, respectively. These results demonstrate that BitMTP accelerates ultra-low-bit LLM generation across backbones, model sizes, \mbox{and GPU types.} \mbox{The source code is released at \codeurl.}
\end{abstract}

\section{Introduction}
\label{sec:intro}
The deployment of large language models (LLMs) is constrained by both model footprint and the serial decoding latency of autoregressive (AR) next-token prediction (NTP)~\citep{ye2025flashinfer,kwon2023efficient,miao2024specinfer}. To resolve this limit, ultra-low-bit models such as BitNet~b1.58 reduce weight storage, memory traffic, and linear-layer execution cost through ternary weights~\citep{wang2025bitnet,ma2025bitnet,team2025minicpm4}, while multi-token prediction (MTP) methods target the sequential bottleneck by predicting multiple future tokens in one pass~\citep{li2024eagle,cai2024medusa,liu2024kangaroo,liu2025sequential,zhang2024draft,samragh2025your,liu2024deepseek}. Despite their complementarity, direct combination of ultra-low-bit models and block-parallel MTP may not accelerate decoding at comparable quality: masked block prediction omits NTP's intermediate token inputs, calling for adaptation to preserve quality, while wider computation and acceptance-dependent causal key--value (KV) updates can offset the latency savings from fewer sequential decoding steps.

\begin{figure}[t]
\centering
\includegraphics[width=\linewidth]{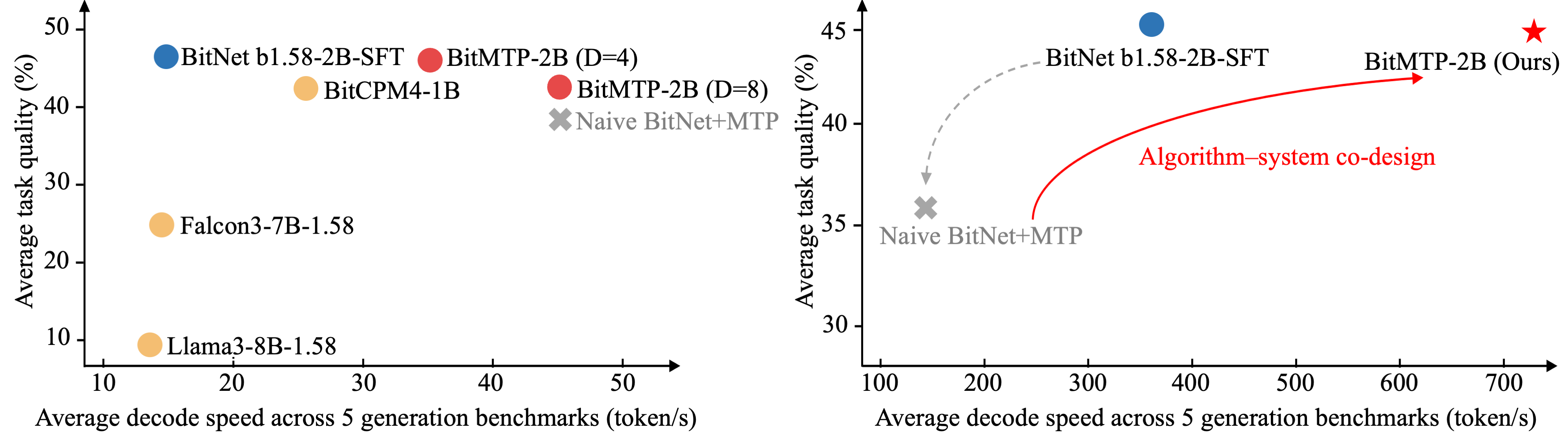}
\caption{Five-task mean generation quality (\%) and decode throughput (tokens/s) for BitNet~b1.58-2B-SFT AR, Naive BitNet$+$MTP, and the proposed BitMTP on NVIDIA H20. The left panel shows the results by using BF16 reference execution, while the right panel uses low-bit W2A8 execution (2-bit weights and 8-bit activations). We see that the naive combination on BF16 improves throughput, but it loses both throughput and quality in the low-bit W2A8 setting. In contrast, our BitMTP substantially increases throughput while retaining \mbox{task quality comparable to SFT AR.}}
\label{fig:bitmtp_motivation}
\end{figure}

To better illustrate the above-mentioned problem, we apply the block-parallel prediction mechanism~\citep{liu2025sequential} to BitNet~b1.58-2B~\citep{ma2025bitnet} and name the fine-tuned model `Naive BitNet$+$MTP'. On NVIDIA H20, we compare its five-task mean quality and decode throughput with the supervised fine-tuning (SFT) AR baseline of BitNet~b1.58-2B. The left panel of Figure~\ref{fig:bitmtp_motivation} shows that the naive combination improves throughput on BF16. However, we see in the right panel of Figure~\ref{fig:bitmtp_motivation} that it loses both throughput and quality in the low-bit W2A8 setting (2-bit weights and 8-bit activations), which shows that fewer decoding steps alone cannot ensure faster low-bit generation. More generally, an MTP decoder's throughput depends on iteration cost and the number of committed tokens. First, a block decoder must execute a wider computation and maintain causal KV states for continued generation. Second, iteration cost is amortized over committed tokens, rather than all predicted tokens. The accepted prefix and decoding policy therefore determine the committed-token count, pending causal KV work, and subsequent replay input. Verifier-based policies additionally require a verification replay. We call the resulting changes in replay inputs, causal KV work, and verifier invocation `accepted-prefix dynamics'. Efficiently executing these transitions turns parallel proposals into \mbox{multi-token speedups on ultra-low-bit LLMs.}

To achieve faster decoding at comparable quality on ultra-low-bit LLM backbones, we propose \textbf{BitMTP}, an algorithm--system co-design framework. At the training level, a joint NTP$+$MTP objective equips the model with mask-conditioned block-causal drafting and an AR path. At the decoding level, confidence-only decoding, speculative-only decoding, and hybrid decoding provide different quality--throughput operating points without using separate draft and verifier models. At the runtime level, three mechanisms are used to reduce block computation and state-update costs. M1 specializes packed low-bit execution by replay width and uses role-dependent output projections. M2 uses a finite-state CUDA Graph atlas to replay reachable pending-prefix states over shared KV storage. M3 prepares verifier inputs before the host branch decision in hybrid decoding. Accepted-prefix alignment keeps the committed prefix, materialized causal KV boundary, and CUDA Graph replay state consistent across decoding and runtime. This supports captured-graph reuse over shared KV history as accepted prefixes change, reducing overhead of multi-token decoding. As shown in Figure~\ref{fig:bitmtp_motivation}, BitMTP achieves substantially higher decode throughput at comparable task quality to the SFT AR baseline. \mbox{Our main contributions are summarized below:}
\begin{itemize}[leftmargin=*, topsep=1pt, itemsep=1pt]
\item \textbf{First end-to-end MTP framework for ultra-low-bit LLMs.} To our knowledge, BitMTP is the first framework to co-design MTP training, decoding, and runtime execution for ultra-low-bit LLMs. Accepted-prefix alignment coordinates decoding decisions with causal KV materialization and graph replay. Captured graphs reuse shared causal KV history across changing accepted prefixes, reducing full-history copying overhead and \mbox{supporting faster low-bit multi-token decoding.}
\item \textbf{Joint training and flexible decoding.} BitMTP combines a joint NTP$+$MTP objective with confidence-only decoding, speculative-only decoding, and hybrid decoding. It thereby supports an AR path and mask-conditioned block-causal drafting without a separate draft--verifier pair. The three decoding policies provide different \mbox{quality--throughput operating points.}
\item \textbf{Replayable runtime for dynamic MTP states.} BitMTP introduces a three-part runtime that reconciles accepted-prefix-dependent execution with fixed-shape CUDA Graph replay. Width-specialized low-bit execution, shared-KV graph replay, and hoisted verifier preparation control the kernel, memory, and scheduling overheads exposed by MTP \mbox{on low-bit LLM backbones.}
\item \textbf{Validation across models and hardware types.} Experiments on BitNet~b1.58-2B and Bonsai models from 1.7B to 4B demonstrate that BitMTP achieves mean decode speedups of up to $2.4\times$ on H20 and $2.0\times$ on RTX~3090. It also reaches a $3.3\times$ MBPP+ speedup on Jetson Orin, demonstrating generalization across backbones, scales, \mbox{and GPU types.}
\end{itemize}

\section{Related Work}
\label{sec:related_work}

\textbf{Low-bit LLMs.} Low-bit LLMs can be obtained by quantizing a pretrained full-precision model or pretraining directly in the low-bit regime. Post-training quantization applies quantization after full-precision training~\citep{xiao2023smoothquant,lin2024awq}, whereas quantization-aware training incorporates the target quantizer during model optimization~\citep{liu2024llm,team2025minicpm4}. For example, Bonsai applies quantization-aware training to pretrained Qwen3 backbones to obtain ternary models~\citep{prismml2026bonsai}. BitNet and BitNet~b1.58 instead use native low-bit pretraining~\citep{ma2024era,wang2025bitnet,ma2025bitnet}. Although low-bit representations reduce weight storage and linear-layer execution cost, token-by-token AR generation still requires a separate model pass for each output token, limiting further throughput gains. Our BitMTP addresses this sequential bottleneck by predicting blocks and committing variable-length continuations on the low-bit backbone.

\textbf{Parallel drafting and blockwise generation.} Recent methods reduce sequential decoding steps through auxiliary heads, learned drafters, or block-parallel prediction~\citep{gloeckle2024better,cai2024medusa,li2024eagle,zhang2024draft}. Medusa adds decoding heads and verifies alternative continuations with tree attention~\citep{cai2024medusa}. EAGLE predicts future hidden states, EAGLE-2 builds context-dependent draft trees, and EAGLE-3 predicts tokens from features fused across target layers~\citep{li2024eagle,li2024eagle2,li2025eagle}. Training approaches include future-summary supervision~\citep{mahajan2025beyond}, non-adjacent targets~\citep{liu2025mtp}, and self-distillation~\citep{kirchenbauer2026multi}. DeepSeek-V3 incorporates MTP during pretraining~\citep{liu2024deepseek}. Masked-input adaptation with gated LoRA and consistency losses exposes a pretrained model's MTP potential~\citep{samragh2025your}. Other methods reuse target-model components or adapt draft structures and budgets~\citep{liu2024kangaroo,zhang2024draft,kim2025fractalllm,chen2024sequoia,hou2025banditspec}. DFlash generates blocks with a target-feature-conditioned diffusion drafter before speculative verification~\citep{chen2026dflash}. Sequential Diffusion Language Models (SDLM) predict masked token blocks and use confidence to commit consecutive prefixes~\citep{liu2025sequential}.
In addition, QSpec~\citep{zhao2025qspec} shares 4-bit weights and a KV cache between W4A4 drafting and W4A16 verification. ML-SpecQD~\citep{georganas2025ml} pairs an MXFP4 draft with a BF16 target and recursively accelerates drafting with a smaller model. QuantSpec~\citep{tiwari2025quantspec} combines hierarchical KV quantization with optional 4-bit draft MLP weights. Although these quantized draft--verify methods combine quantization with speculative decoding to reduce drafting and memory costs, sequential proposal dependencies and recurring verification still constrain further acceleration. Our BitMTP combines block-parallel proposals, conditional verification in hybrid decoding, and accepted-prefix-aligned low-bit execution to \mbox{reduce work per committed token.}

\textbf{Inference runtimes for efficient generation.} The CPU inference system of BitNet provides specialized kernels for ternary execution~\citep{wang2025bitnetcpp}. GPU systems such as Atom and QServe co-design low-bit quantization and serving optimizations~\citep{zhao2024atom,lin2025qserve}. Other work optimizes attention and decode kernels~\citep{ye2025flashinfer,dao2022flashattention,shah2024flashattention,hong2024flashdecoding++}. KV-cache techniques address allocation, compression, and quantization~\citep{prabhu2025vattention,lee2024infinigen,liu2024minicache,hooper2024kvquant,cheng2025qaq}. Serving systems improve request scheduling and execution~\citep{sun2024llumnix,agrawal2024taming,zheng2024sglang,zhong2024distserve}. For block generation on an ultra-low-bit LLM, the accepted prefix and verification branch determine causal KV materialization and the next replay input. Kernel, cache, and scheduling optimizations alone do not provide a coordinated end-to-end acceleration path for these transitions. Our proposed BitMTP provides this path through width-specialized packed operators, captured graphs over shared KV storage, and preparation of verifier inputs \mbox{before the hybrid branch decision.}

\section{The BitMTP Framework}
\label{sec:method}

\subsection{Method Overview}
\label{sec:accepted_prefix}
\begin{figure}[t]
\centering
\includegraphics[width=\linewidth]{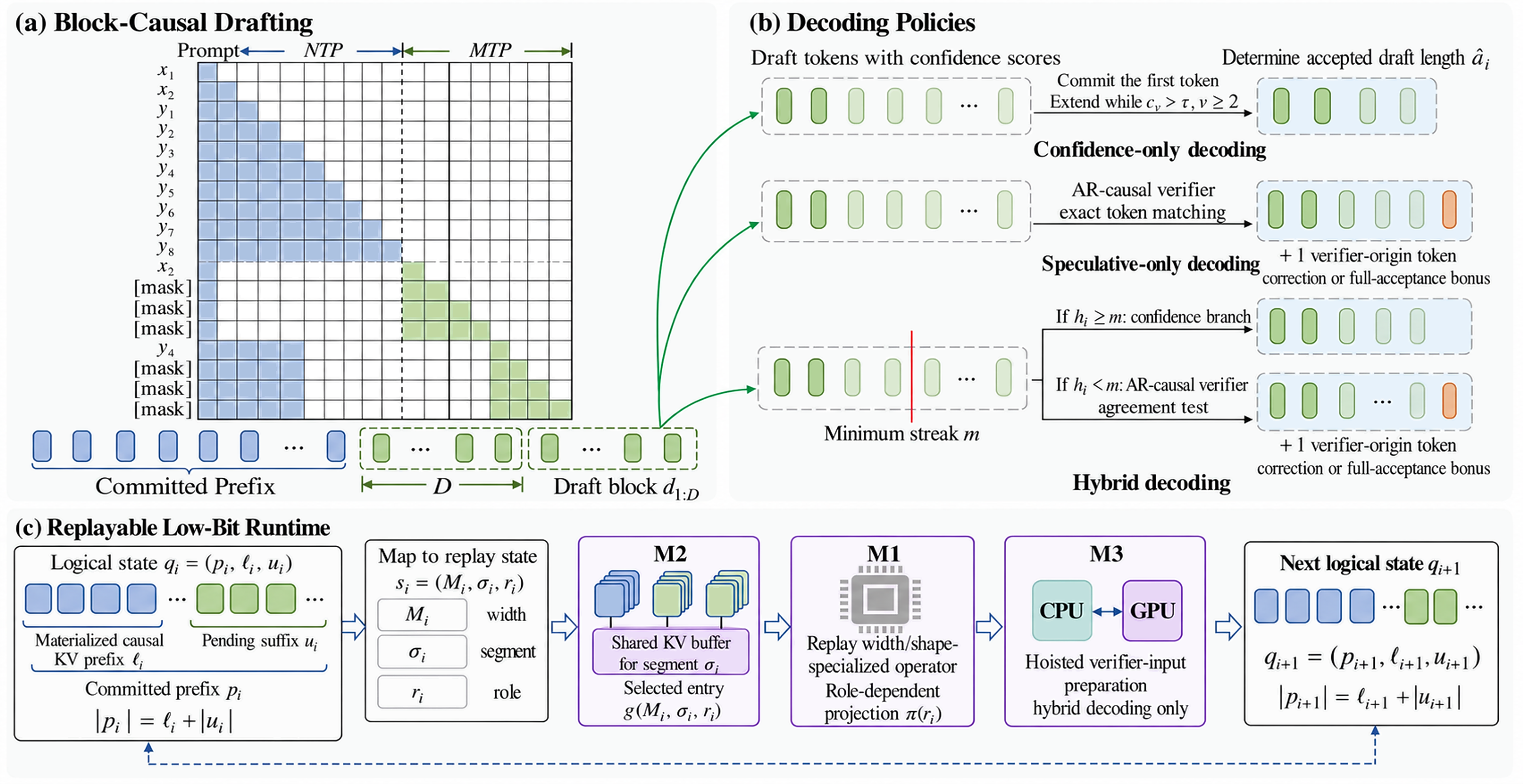}
\caption{The BitMTP framework. (a)~Joint NTP$+$MTP training supports mask-conditioned block-causal drafting and an AR path in one model. (b)~Confidence-only decoding commits a confidence-selected prefix. Speculative-only decoding verifies each block, while hybrid decoding verifies when the confidence prefix is too short. Verifier-based steps add a correction or bonus token. (c)~The runtime executes accepted-prefix transitions with width-specialized low-bit operators and captured CUDA Graphs. Graphs reuse causal KV history through shared storage. In hybrid decoding, verifier inputs are prepared before the host branch decision.}
\label{fig:bitmtp_overview}
\end{figure}

As illustrated in Figure~\ref{fig:bitmtp_overview}, BitMTP enables low-bit multi-token generation through coordinated training, decoding, and runtime components. Joint training equips the model with both AR generation and mask-conditioned block-causal drafting. The decoding component uses confidence-only decoding, speculative-only decoding, or hybrid decoding to select the committed prefix. The runtime executes the selected transition with fixed-shape low-bit CUDA Graph replay. These components satisfy the same consistency requirement: the committed prefix, materialized causal KV boundary, and CUDA Graph replay state must remain mutually consistent among decoding iterations, which \mbox{we call `accepted-prefix alignment'}.

Each draft block proposes $D$ future tokens. At iteration $i$, decoding tracks $q_i=(p_i,\ell_i,u_i)$: the committed prefix $p_i$, its materialized causal KV length $\ell_i$, and pending suffix $u_i$. The prefix includes the prompt and emitted tokens. Accepted-prefix alignment imposes the following length relation:
\begin{equation}
|p_i|=\ell_i+n_i,
\label{eq:prefix_invariant}
\end{equation}
where $n_i=|u_i|$ counts committed tokens awaiting causal KV.
During initialization, prefill processes the prompt and builds its causal KV prefix. Since tokens emitted from the initialization logits have not yet been processed as causal inputs, they form the pending suffix. Later, $u_i$ will contain confidence-committed drafts or a verifier-origin token, and the next draft replay computes their causal KV \mbox{together with a new block.}

The runtime maps $q_i$ and the selected decoding policy to a replay state $s_i=(M_i,\sigma_i,r_i)$. The width $M_i$ covers the pending suffix and the next block computation. The segment $\sigma_i$ selects the active KV capacity, and $r_i$ selects the numerical execution role. A verifier step additionally selects a width-$(D{+}1)$ replay with the verifier role. Each draft step computes and materializes the pending-token KV in $q_i$; the next state records committed tokens. Algorithm~\ref{alg:bitmtp_iteration} in \textbf{Appendix~\ref{app:method_details}} \mbox{summarizes one hybrid decoding iteration.}

\subsection{Block-Causal Drafting and Training}
\label{sec:model_side}
For a training sequence $y$ anchored at $s_b$, BitMTP constructs the block input and targets:
\begin{equation}
X_b=[y_{s_b},m_{\mathrm{mask}},\ldots,m_{\mathrm{mask}}],
\qquad
Y_b=[y_{s_b+1},\ldots,y_{s_b+D}],
\label{eq:block_draft}
\end{equation}
where $y$ includes the prompt and the response, and $m_{\mathrm{mask}}$ denotes the mask input. The observed anchor occupies the first input slot, and mask embeddings occupy the remaining $D{-}1$ slots. The targets in $Y_b$ are the $D$ ordered future tokens. Attention within the block is causal, as illustrated in Figure~\ref{fig:bitmtp_overview}(a). Each prediction can attend to the anchor and the mask slots preceding its target position. Because the block inputs are constructed before the forward pass, the model produces \mbox{the whole draft block in parallel.}

We retain causal next-token supervision and combine it with block prediction. The summed losses are then normalized over all supervised targets:
\begin{equation}
\begin{aligned}
\mathcal{L}_{\mathrm{mtp}}
&=
\sum_{b\in\mathcal{B}}\sum_{u\in\mathcal{U}_b}
-\log p_\theta\!\left(y_{s_b+u}\mid y_{<s_b},X_{b,1:u}\right),
\qquad
\mathcal{L}
=\frac{\mathcal{L}_{\mathrm{ar}}+\mathcal{L}_{\mathrm{mtp}}}
{N_{\mathrm{ar}}+N_{\mathrm{mtp}}},
\end{aligned}
\label{eq:bitmtp_loss}
\end{equation}
where $\theta$ denotes the model parameters. The set $\mathcal{B}$ contains training blocks, and $\mathcal{U}_b\subseteq\{1,\ldots,D\}$ contains the valid supervised slots of block $b$. The slice $X_{b,1:u}$ comprises its first $u$ input slots. The AR loss $\mathcal{L}_{\mathrm{ar}}$ sums causal next-token negative log-likelihoods, while $N_{\mathrm{ar}}$ and $N_{\mathrm{mtp}}$ count supervised AR and block targets. The AR term supports causal generation, while the MTP term learns block-causal proposals. The resulting model performs \mbox{both drafting and AR verification.}

\subsection{Decoding Policies}
\label{sec:strategy}
BitMTP provides confidence-only decoding, speculative-only decoding, and hybrid decoding, as shown in Figure~\ref{fig:bitmtp_overview}(b). Each policy operates on a draft block $d_{1:D}$ and determines the accepted draft length $\hat a_i$. A confidence step forms an output block of $\hat a_i$ draft tokens. A verifier step adds one verifier-origin token, giving $\hat a_i+1$ \mbox{tokens in the block.}

\textbf{Confidence-only decoding.}
With draft-token softmax probability $c_u$ and threshold $\tau$, the decoder always commits $d_1$ and extends the prefix through consecutive confident drafts:
\begin{equation}
h_i=1+\sum_{u=2}^{D}\prod_{v=2}^{u}\mathbf{1}[c_v>\tau].
\label{eq:confidence_accept}
\end{equation}
The $h_i$-token confidence streak is committed without verification, giving $\hat a_i=h_i$.

\textbf{Speculative-only decoding.}
The same model processes $[\text{anchor},d_1,\ldots,d_D]$ under an AR-causal mask, where the anchor is the last committed token. It produces verifier tokens $\tilde d_{1:D+1}$. Verification selects the~longest~matching~prefix:
\begin{equation}
\hat a_i=
\max\left\{k\in\{0,\ldots,D\}:
d_u=\tilde d_u\ \text{for every}\ u\le k\right\}.
\label{eq:exact_verify}
\end{equation}
The decoder forms an output block containing $d_{1:\hat a_i}$, followed by $b_i=\tilde d_{\hat a_i+1}$. The verifier-origin token $b_i$ corrects the first rejected draft position or supplies a bonus token after full acceptance. Speculative-only decoding preserves the greedy continuation defined by \mbox{the AR-causal verifier.}

\textbf{Hybrid decoding.}
Hybrid decoding uses a minimum confidence-prefix length $m$. It sets $\hat a_i=h_i$ and commits the confidence prefix when $h_i\ge m$. Otherwise, it invokes the AR-causal verifier to determine the accepted prefix and verifier-origin token.

\subsection{Replayable Low-Bit Runtime}
\label{sec:runtime}
\begin{figure}[t]
\centering
\includegraphics[width=\linewidth]{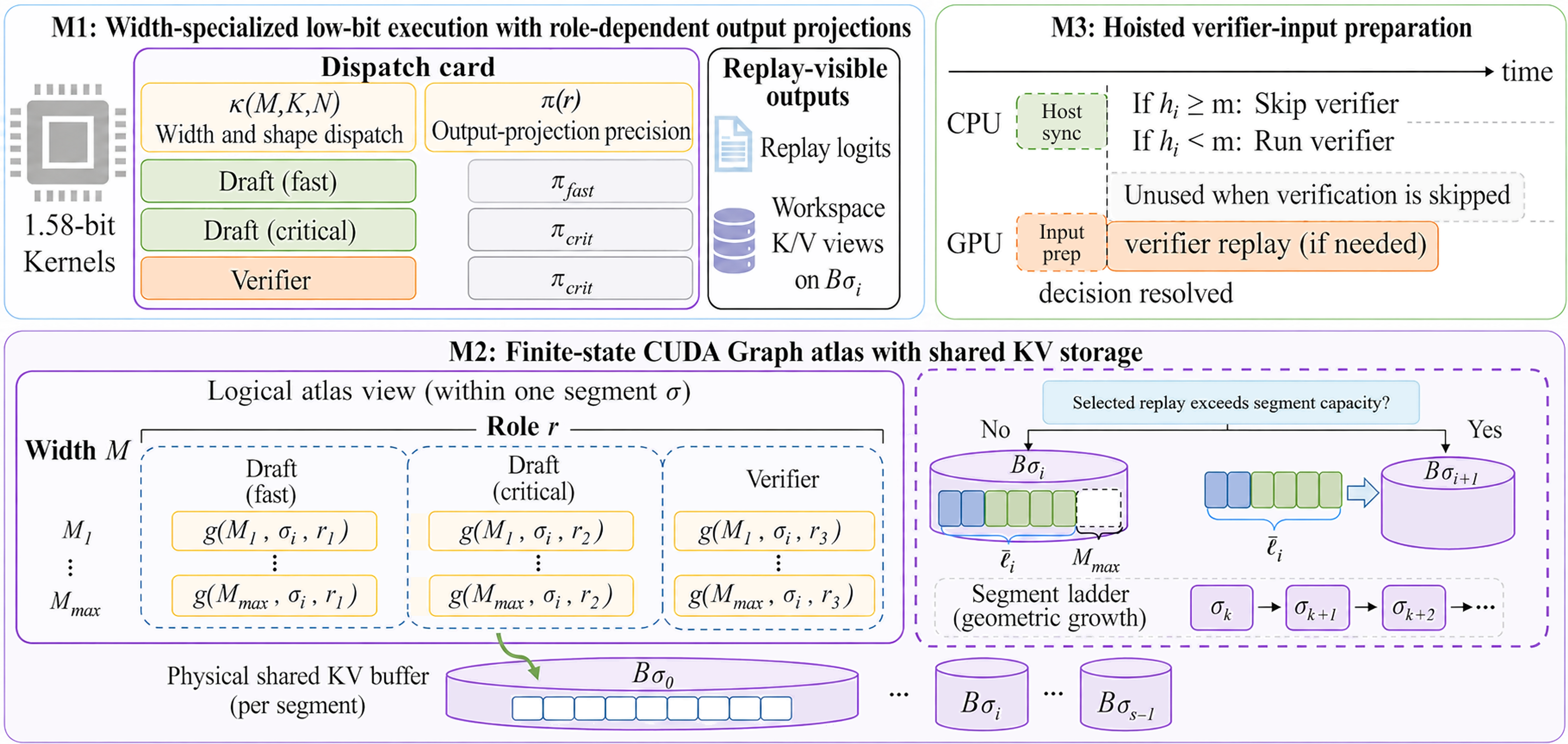}
\caption{Runtime mechanisms M1--M3. M1 selects packed operators by replay width $M$ and input/output dimensions $(K,N)$, while numerical role $r$ selects the output projection $\pi(r)$. M2 indexes captured graphs by $(M,\sigma,r)$ and shares one KV buffer $B_\sigma$ across widths and roles within each capacity segment. M3 moves verifier-input preparation ahead of the hybrid branch decision.}
\label{fig:bitmtp_runtime_state}
\end{figure}

Figure~\ref{fig:bitmtp_runtime_state} shows how M1--M3 implement a decoding transition. At each step, BitMTP selects a replay state $s_i=(M_i,\sigma_i,r_i)$ specifying the replay width, KV-capacity segment, and numerical role. M1 selects the packed operator and output projection, M2 replays the corresponding graph over shared causal history, and M3 moves verifier-input preparation \mbox{ahead of the hybrid branch decision}.

After a draft replay, the runtime copies the $n_i$ pending-token KV rows into history. The selected decoding policy then updates the state as follows:
\begin{equation}
\begin{aligned}
\bar\ell_i &= \ell_i+n_i, \\
q_{i+1}
&=
\begin{cases}
\bigl(p_i\!\cdot\!d_{1:\hat a_i},\,\bar\ell_i,\,d_{1:\hat a_i}\bigr),
& I_{\mathrm{verify}}(i)=0,\\
\bigl(p_i\!\cdot\!d_{1:\hat a_i}\!\cdot\!b_i,\,\bar\ell_i+\hat a_i,\,[b_i]\bigr),
& I_{\mathrm{verify}}(i)=1,
\end{cases}
\end{aligned}
\label{eq:kv_evolution}
\end{equation}
where $\bar\ell_i$ is the materialized KV length after processing the pending suffix, and $b_i$ is the verifier-origin token from Section~\ref{sec:strategy}. The indicator $I_{\mathrm{verify}}(i)$ is $0$ for confidence-only decoding, $1$ for speculative-only decoding, and $\mathbf{1}[h_i<m]$ for hybrid decoding. Without verification, the committed draft prefix remains pending. With verification, its accepted KV rows are materialized and only $b_i$ remains pending. Both branches preserve $|p_{i+1}|=\ell_{i+1}+|u_{i+1}|$.

\textbf{M1: Width-specialized low-bit execution with role-dependent output projections.}
M1 specializes packed low-bit execution for the replay widths used by AR generation, block drafting, and verification. As shown in the top-left panel of Figure~\ref{fig:bitmtp_runtime_state}, M1 selects a packed operator by $(M,K,N)$ and an output projection by numerical role $r$. The dispatch is:
\begin{equation}
\kappa:(M,K,N)\longmapsto\mathcal{K}_{M,K,N},
\qquad
\pi(r)=
\begin{cases}
\pi_{\mathrm{fast}},&r\in\mathcal{R}_{\mathrm{fast}},\\
\pi_{\mathrm{crit}},&r\in\mathcal{R}_{\mathrm{crit}}\cup\mathcal{R}_{\mathrm{ver}}
\end{cases},
\label{eq:rws_primitive}
\end{equation}
where $\mathcal{K}_{M,K,N}$ is the selected packed operator, and $K,N$ are its input and output feature dimensions. The set $\mathcal{R}_{\mathrm{fast}}$ contains fast drafting roles, $\mathcal{R}_{\mathrm{crit}}$ contains verification-critical drafting and AR roles, and $\mathcal{R}_{\mathrm{ver}}$ contains verifier roles. Drafting selects its projection through $r$, while AR generation and verification use $\pi_{\mathrm{crit}}$. All roles share the packed low-bit transformer body and project only the token rows required by the replay.

\textbf{M2: Finite-state CUDA Graph atlas with shared KV storage.}
Accepted-prefix dynamics change both replay width and the materialized KV boundary, whereas CUDA Graph replay requires fixed shapes and addresses. As shown in Figure~\ref{fig:bitmtp_runtime_state}, M2 supports these changes with shared segment storage and captures a graph set $\mathcal{G}$ covering reachable transitions from a candidate graph set:
\begin{equation}
\mathcal{G}_{\mathrm{cand}}=\big\{g_{M,\sigma,r}:M\!\in\!\mathcal{M},\ \sigma\!\in\!\Sigma,\ r\!\in\!\mathcal{R}\big\},
\qquad
\mathcal{G}\subseteq\mathcal{G}_{\mathrm{cand}},
\label{eq:graph_atlas}
\end{equation}
where $g_{M,\sigma,r}$ denotes a graph with replay key $(M,\sigma,r)$. The sets $\mathcal{M}$, $\Sigma$, and $\mathcal{R}$ contain supported replay widths, ordered KV-capacity segments, and M1 execution roles, respectively. The captured set covers reachable replay states, including width-$(D{+}1)$ verification.

Within segment $\sigma$, all graphs use fixed-shape views of one physical KV buffer $B_\sigma$, with $|\sigma|$ history slots and $M_{\max}=\max\mathcal{M}$ workspace slots. Width and role switches therefore preserve materialized history. After drafting, the runtime copies the pending-token KV rows into history; after verification, it promotes accepted draft rows, excluding the repeated anchor. These operations copy consecutive token rows within the segment. When a transition requires a larger capacity segment, the runtime copies the materialized history into the new buffer. Graphs within each segment thus reuse one causal KV history while implementing Eq.~\ref{eq:kv_evolution}.

Shared KV storage is proportional to $\sum_{\sigma\in\Sigma}(|\sigma|+M_{\max})$ for a fixed model. With fixed replay widths, geometrically increasing capacities cover history length $\ell_{\max}$ using $O(\log\ell_{\max})$ segments and $O(\ell_{\max})$ KV slots. \textbf{Appendix~\ref{app:method_details}} details the replay masks and KV copies.

\textbf{M3: Hoisted verifier-input preparation.}
M3 moves verifier-input preparation ahead of the branch decision in hybrid decoding, as shown in Figure~\ref{fig:bitmtp_runtime_state}. The draft block, committed anchor, and materialized causal KV boundary determine the verifier inputs before the host reads $h_i$. After drafting, the runtime enqueues preparation of the width-$(D{+}1)$ verifier token buffer, positions, and AR-causal mask on the GPU stream, then obtains $h_i$ on the host. Preparation is issued in each hybrid iteration, while verifier replay occurs only when $h_i<m$. The selected verifier graph therefore consumes the prepared inputs without a separate input-preparation stage after the branch decision.

\section{Experimental Evaluation}
\label{sec:experiments}

\subsection{Experimental Setup}
\label{sec:exp_setup}
\textbf{Models, training, and decoding.}
We evaluate three ultra-low-bit LLM backbones: BitNet~b1.58-2B~\citep{ma2025bitnet}, Bonsai-1.7B, and Bonsai-4B~\citep{prismml2026bonsai}. For each backbone, additional supervised fine-tuning produces the SFT model used as the AR baseline. The SFT and MTP variants use the same processed corpus, with MTP using the joint objective in Eq.~\ref{eq:bitmtp_loss}. Generation uses greedy decoding. Primary confidence-only decoding and hybrid decoding use $\tau{=}0.9$. Hybrid decoding uses the model's AR-causal verifier to determine the accepted prefix when verification is required. \textbf{Appendix~\ref{app:training_data}} summarizes the training corpus and reports one-epoch \mbox{training cost on Bonsai-1.7B.}

\textbf{Hardware and low-bit backends.}
Server, consumer, and edge tests use NVIDIA H20, RTX~3090, and Jetson Orin NX, respectively. The PyTorch/CUDA runtime uses packed W2A8 for decode-stage transformer linear layers, with INT2 weights for BitNet and group-128 ternary weights for Bonsai. H20 enables FP8 draft projections for the fast drafting roles in M1. The FP8 GEMM backend does not support the Ampere GPUs in RTX~3090 and Orin, so these platforms use non-FP8 draft projections. Orin runs under \texttt{MAXN\_SUPER} with locked clocks. AR generation and verifier replay use verification-critical projections. \mbox{Other baseline systems use their released runtimes.}

\textbf{Generation benchmarks.}
The generation suite covers IFEval for instruction following, GSM8K and MATH-500 for reasoning, and HumanEval+ and MBPP+ for code generation. We report strict instruction-following accuracy, exact match for reasoning, and EvalPlus pass@1 for code.

\textbf{Metrics and model capability evaluation.}
Decode throughput divides timed output tokens by decode-loop time measured after prefill and request initialization. Task means are unweighted across the five generation benchmarks, and speedups use the matched SFT AR route on the same device and backend. The eleven-task capability evaluation uses \texttt{lm-eval} with \mbox{full benchmark sets and standard prompts.}

\subsection{Generation Quality and Throughput}
\label{sec:exp_main}
Table~\ref{tab:gen_bench_main} compares compact BF16 models~\citep{hu2024minicpm,qwen2.5,yang2025qwen3}, quantized and low-bit AR models, and learned draft--verify systems with BitMTP on H20. The matched SFT AR routes establish the acceleration obtained beyond packed low-bit execution. At $D{=}8$, hybrid decoding achieves five-task mean speedups of $2.2\times$ on BitNet, $2.4\times$ on Bonsai-1.7B, and $2.1\times$ on Bonsai-4B. The Bonsai mean scores remain close to SFT: $47.91$ versus $47.36$ and $63.43$ versus $63.55$. The compact-model comparisons further place this quality in context. With hybrid decoding, BitMTP--Bonsai-1.7B approaches BF16 Qwen2.5-1.5B's $49.23$ and exceeds its AWQ variant's $41.79$, while BitMTP--Bonsai-4B exceeds Llama-3.2-3B's $57.41$ and Qwen3-1.7B's $58.23$. BitMTP therefore supports multi-token acceleration on both smaller deployment \mbox{models and stronger backbones.}

\suppressfloats[t]
\begin{table}[!t]
\centering
\caption{Generation quality (Sc., \%) and decode throughput (Rate, tokens/s) on NVIDIA H20. Dark-red values in parentheses give speedups over AR baselines. Average is the unweighted five-task mean. W/A denotes weight/activation precision. Conf., Spec., and Hybr. denote confidence-only, speculative-only, \mbox{and hybrid decoding, respectively.}}
\label{tab:gen_bench_main}
\footnotesize
\setlength{\tabcolsep}{1.5pt}
\renewcommand{\arraystretch}{1.02}
\resizebox{\linewidth}{!}{%
\begin{tabular}{ccc*{5}{cc}cc}
\toprule
\multirowcell{2}{Model} & \multirowcell{2}{W/A} & \multirowcell{2}{Type} &
\multicolumn{2}{c}{GSM8K} & \multicolumn{2}{c}{MATH-500} &
\multicolumn{2}{c}{IFEval} & \multicolumn{2}{c}{HumanEval+} &
\multicolumn{2}{c}{MBPP+} & \multicolumn{2}{c}{Average}\\
\cmidrule(lr){4-5}\cmidrule(lr){6-7}\cmidrule(lr){8-9}\cmidrule(lr){10-11}\cmidrule(lr){12-13}\cmidrule(lr){14-15}
& & & Sc. & Rate & Sc. & Rate & Sc. & Rate & Sc. & Rate & Sc. & Rate & Sc. & Rate\\
\midrule
\multicolumn{15}{c}{\textit{Compact, quantized, and speculative baselines}}\\
\midrule
MiniCPM-2B & BF16/BF16 & AR & 47.84 & 273.83 & 11.00 & 301.18 & 43.65 & 306.92 & 40.24 & 290.12 & 43.65 & 252.08 & 37.28 & 284.82\\
Qwen2.5-1.5B & BF16/BF16 & AR & 57.54 & 441.38 & 49.00 & 470.88 & 51.08 & 467.00 & 37.20 & 439.50 & 51.32 & 395.61 & 49.23 & 442.88\\
Llama-3.2-3B & BF16/BF16 & AR & 66.79 & 274.71 & 45.80 & 303.84 & 78.90 & 303.35 & 44.51 & 284.95 & 51.06 & 256.08 & 57.41 & 284.59\\
Qwen3-1.7B & BF16/BF16 & AR & 68.16 & 452.38 & 67.40 & 454.19 & 75.30 & 457.90 & 37.20 & 456.47 & 43.12 & 464.42 & 58.23 & 457.07\\
\midrule
Qwen2.5-1.5B-AWQ & INT4/FP16 & AR & 50.49 & 266.26 & 37.80 & 268.65 & 47.12 & 270.05 & 29.88 & 268.86 & 43.65 & 267.39 & 41.79 & 268.24\\
Llama3-8B-1.58$^{\dagger}$ & 1.58/BF16 & AR & 3.94 & 13.77 & 1.00 & 13.79 & 30.70 & 13.83 & 0.00 & 14.06 & 0.79 & 14.03 & 7.29 & 13.90\\
Falcon3-7B-1.58bit$^{\dagger}$ & 1.58/BF16 & AR & 1.67 & 14.99 & 18.40 & 14.98 & 43.76 & 13.84 & 35.98 & 15.24 & 23.02 & 15.71 & 24.56 & 14.95\\
BitCPM4-1B$^{\dagger}$ & 1.58/BF16 & AR & 41.85 & 26.66 & 29.80 & 22.81 & 72.18 & 29.32 & 17.07 & 27.95 & 52.12 & 27.88 & 42.60 & 26.92\\
\midrule
\multicolumn{15}{c}{\textit{BitNet~b1.58-2B family: packed W2A8}}\\
\midrule
BitNet b1.58-2B & 1.58/INT8 & AR & 44.28 & 363.53 & 40.80 & 373.12 & 62.11 & 360.55 & 34.76 & 355.99 & 34.92 & 378.73 & 43.37 & 366.38\\
BitNet b1.58-2B-SFT & 1.58/INT8 & AR & 58.00 & 364.97 & 35.80 & 364.97 & 55.76 & 368.73 & 40.24 & 393.65 & 39.95 & 374.78 & 45.95 & 373.42\\
\cmidrule(lr){2-15}
\multirow{6}{*}{\makecell{BitMTP--BitNet b1.58-2B\\($D{=}4$)}} & \multirow{6}{*}{1.58/INT8} & \multirowcell{2}{MTP\\(Conf.)} & 61.11 & 691.59 & 32.80 & 614.13 & 54.80 & 467.65 & 35.37 & 698.26 & 32.54 & 682.60 & 43.32 & 630.84\\[-1.5pt]
 &  &  &  & \submetric{1.9$\times$} &  & \submetric{1.7$\times$} &  & \submetric{1.3$\times$} &  & \submetric{1.8$\times$} &  & \submetric{1.8$\times$} &  & \submetric{1.7$\times$}\\[0.5pt]
 &  & \multirowcell{2}{MTP\\(Spec.)} & 65.05 & 662.78 & 37.20 & 640.38 & 54.44 & 535.81 & 34.76 & 658.55 & 34.39 & 671.77 & 45.17 & 633.86\\[-1.5pt]
 &  &  &  & \submetric{1.8$\times$} &  & \submetric{1.8$\times$} &  & \submetric{1.5$\times$} &  & \submetric{1.7$\times$} &  & \submetric{1.8$\times$} &  & \submetric{1.7$\times$}\\[0.5pt]
 &  & \multirowcell{2}{MTP\\(Hybr.)} & 63.08 & 761.42 & 32.60 & 707.11 & 55.64 & 580.35 & 33.54 & 768.71 & 34.92 & 793.01 & 43.95 & 722.12\\[-1.5pt]
 &  &  &  & \submetric{2.1$\times$} &  & \submetric{1.9$\times$} &  & \submetric{1.6$\times$} &  & \submetric{2.0$\times$} &  & \submetric{2.1$\times$} &  & \submetric{1.9$\times$}\\[0.5pt]
\cmidrule(lr){2-15}
\multirow{6}{*}{\makecell{BitMTP--BitNet b1.58-2B\\($D{=}8$)}} & \multirow{6}{*}{1.58/INT8} & \multirowcell{2}{MTP\\(Conf.)} & 56.48 & 647.49 & 29.60 & 539.27 & 53.36 & 409.84 & 32.32 & 691.87 & 31.75 & 728.68 & 40.70 & 603.43\\[-1.5pt]
 &  &  &  & \submetric{1.8$\times$} &  & \submetric{1.5$\times$} &  & \submetric{1.1$\times$} &  & \submetric{1.8$\times$} &  & \submetric{1.9$\times$} &  & \submetric{1.6$\times$}\\[0.5pt]
 &  & \multirowcell{2}{MTP\\(Spec.)} & 60.35 & 718.60 & 32.60 & 698.14 & 55.28 & 503.21 & 34.15 & 772.56 & 32.80 & 790.60 & 43.04 & 696.62\\[-1.5pt]
 &  &  &  & \submetric{2.0$\times$} &  & \submetric{1.9$\times$} &  & \submetric{1.4$\times$} &  & \submetric{2.0$\times$} &  & \submetric{2.1$\times$} &  & \submetric{1.9$\times$}\\[0.5pt]
 &  & \multirowcell{2}{MTP\\(Hybr.)} & 60.35 & 819.42 & 33.60 & 776.99 & 54.56 & 543.59 & 34.15 & 923.63 & 33.60 & 964.20 & 43.25 & 805.57\\[-1.5pt]
 &  &  &  & \submetric{2.2$\times$} &  & \submetric{2.1$\times$} &  & \submetric{1.5$\times$} &  & \submetric{2.3$\times$} &  & \submetric{2.6$\times$} &  & \submetric{2.2$\times$}\\[0.5pt]
\midrule
\multicolumn{15}{c}{\textit{Bonsai-1.7B family: packed W2A8}}\\
\midrule
Bonsai-1.7B & 1.58/INT8 & AR & 64.14 & 469.36 & 46.20 & 449.62 & 78.78 & 465.63 & 41.46 & 475.39 & 20.90 & 473.44 & 50.30 & 466.69\\
Bonsai-1.7B-SFT & 1.58/INT8 & AR & 59.67 & 472.99 & 41.20 & 447.08 & 68.59 & 473.98 & 31.10 & 477.83 & 36.24 & 474.42 & 47.36 & 469.26\\
\cmidrule(lr){2-15}
\multirow{2}{*}{Bonsai-1.7B $+$ EAGLE-3} & \multirow{2}{*}{W2A8$+$BF16} & \multirow{2}{*}{MTP} & 58.98 & 515.55 & 40.80 & 537.35 & 68.71 & 547.04 & 33.54 & 600.11 & 36.24 & 631.10 & 47.65 & 566.23\\[-1.5pt]
 &  &  &  & \submetric{1.1$\times$} &  & \submetric{1.2$\times$} &  & \submetric{1.2$\times$} &  & \submetric{1.3$\times$} &  & \submetric{1.3$\times$} &  & \submetric{1.2$\times$}\\[0.5pt]
\cmidrule(lr){2-15}
\multirow{6}{*}{\makecell{BitMTP--Bonsai-1.7B\\($D{=}4$)}} & \multirow{6}{*}{1.58/INT8} & \multirowcell{2}{MTP\\(Conf.)} & 54.89 & 701.25 & 41.80 & 837.48 & 64.75 & 648.91 & 32.93 & 943.90 & 36.51 & 847.30 & 46.17 & 795.77\\[-1.5pt]
 &  &  &  & \submetric{1.5$\times$} &  & \submetric{1.9$\times$} &  & \submetric{1.4$\times$} &  & \submetric{2.0$\times$} &  & \submetric{1.8$\times$} &  & \submetric{1.7$\times$}\\[0.5pt]
 &  & \multirowcell{2}{MTP\\(Spec.)} & 57.85 & 795.80 & 44.20 & 853.38 & 66.43 & 714.93 & 32.93 & 888.55 & 38.89 & 867.85 & 48.06 & 824.10\\[-1.5pt]
 &  &  &  & \submetric{1.7$\times$} &  & \submetric{1.9$\times$} &  & \submetric{1.5$\times$} &  & \submetric{1.9$\times$} &  & \submetric{1.8$\times$} &  & \submetric{1.8$\times$}\\[0.5pt]
 &  & \multirowcell{2}{MTP\\(Hybr.)} & 57.47 & 901.35 & 42.20 & 979.79 & 65.35 & 791.36 & 32.32 & 1078.32 & 38.89 & 1010.11 & 47.24 & 952.18\\[-1.5pt]
 &  &  &  & \submetric{1.9$\times$} &  & \submetric{2.2$\times$} &  & \submetric{1.7$\times$} &  & \submetric{2.3$\times$} &  & \submetric{2.1$\times$} &  & \submetric{2.0$\times$}\\[0.5pt]
\cmidrule(lr){2-15}
\multirow{6}{*}{\makecell{BitMTP--Bonsai-1.7B\\($D{=}8$)}} & \multirow{6}{*}{1.58/INT8} & \multirowcell{2}{MTP\\(Conf.)} & 54.21 & 661.80 & 39.60 & 837.24 & 66.19 & 593.09 & 34.15 & 1010.59 & 35.71 & 1040.99 & 45.97 & 828.74\\[-1.5pt]
 &  &  &  & \submetric{1.4$\times$} &  & \submetric{1.9$\times$} &  & \submetric{1.3$\times$} &  & \submetric{2.1$\times$} &  & \submetric{2.2$\times$} &  & \submetric{1.8$\times$}\\[0.5pt]
 &  & \multirowcell{2}{MTP\\(Spec.)} & 57.70 & 907.20 & 41.60 & 1051.18 & 68.11 & 760.12 & 34.76 & 1184.89 & 37.30 & 1237.25 & 47.89 & 1028.13\\[-1.5pt]
 &  &  &  & \submetric{1.9$\times$} &  & \submetric{2.4$\times$} &  & \submetric{1.6$\times$} &  & \submetric{2.5$\times$} &  & \submetric{2.6$\times$} &  & \submetric{2.2$\times$}\\[0.5pt]
 &  & \multirowcell{2}{MTP\\(Hybr.)} & 57.47 & 919.54 & 43.20 & 1112.68 & 68.59 & 801.42 & 33.54 & 1288.33 & 36.77 & 1431.28 & 47.91 & 1110.65\\[-1.5pt]
 &  &  &  & \submetric{1.9$\times$} &  & \submetric{2.5$\times$} &  & \submetric{1.7$\times$} &  & \submetric{2.7$\times$} &  & \submetric{3.0$\times$} &  & \submetric{2.4$\times$}\\[0.5pt]
\midrule
\multicolumn{15}{c}{\textit{Bonsai-4B family: packed W2A8}}\\
\midrule
Bonsai-4B-SFT & 1.58/INT8 & AR & 79.08 & 278.04 & 61.20 & 273.46 & 76.62 & 277.11 & 52.44 & 278.66 & 48.41 & 279.59 & 63.55 & 277.37\\
\cmidrule(lr){2-15}
\multirow{6}{*}{\makecell{BitMTP--Bonsai-4B\\($D{=}8$)}} & \multirow{6}{*}{1.58/INT8} & \multirowcell{2}{MTP\\(Conf.)} & 76.35 & 347.76 & 55.40 & 473.75 & 74.94 & 281.91 & 48.78 & 545.92 & 53.17 & 509.61 & 61.73 & 431.79\\[-1.5pt]
 &  &  &  & \submetric{1.3$\times$} &  & \submetric{1.7$\times$} &  & \submetric{1.0$\times$} &  & \submetric{2.0$\times$} &  & \submetric{1.8$\times$} &  & \submetric{1.6$\times$}\\[0.5pt]
 &  & \multirowcell{2}{MTP\\(Spec.)} & 78.54 & 504.21 & 62.20 & 564.80 & 78.30 & 365.03 & 50.00 & 612.29 & 51.59 & 603.92 & 64.13 & 530.05\\[-1.5pt]
 &  &  &  & \submetric{1.8$\times$} &  & \submetric{2.1$\times$} &  & \submetric{1.3$\times$} &  & \submetric{2.2$\times$} &  & \submetric{2.2$\times$} &  & \submetric{1.9$\times$}\\[0.5pt]
 &  & \multirowcell{2}{MTP\\(Hybr.)} & 78.17 & 513.27 & 59.60 & 612.41 & 77.34 & 369.50 & 49.39 & 701.86 & 52.65 & 709.18 & 63.43 & 581.24\\[-1.5pt]
 &  &  &  & \submetric{1.8$\times$} &  & \submetric{2.2$\times$} &  & \submetric{1.3$\times$} &  & \submetric{2.5$\times$} &  & \submetric{2.5$\times$} &  & \submetric{2.1$\times$}\\[0.5pt]
\bottomrule
\end{tabular}}

\par\smallskip\noindent\begin{minipage}{\linewidth}\footnotesize
$^{\dagger}$BF16 reference execution.
\end{minipage}
\end{table}

The EAGLE-3~\citep{li2025eagle} comparison examines learned speculative drafting with a packed low-bit target. Bonsai$+$EAGLE-3 reports $566.23$ tokens/s at $47.65$ mean quality, whereas BitMTP hybrid decoding at $D{=}8$ reaches $1110.65$ tokens/s at $47.91$. BitMTP combines block proposals and AR verification in one low-bit backbone rather than adding a BF16 drafter. Hybrid decoding exceeds confidence-only decoding in both mean quality and throughput across all five model--block-size configurations in Table~\ref{tab:gen_bench_main}. On Bonsai-1.7B at $D{=}8$, hybrid decoding and speculative-only decoding score $47.91$ and $47.89$, while hybrid throughput increases from $1028.13$ to $1110.65$ tokens/s. BitNet offers a different choice when quality retention is prioritized: $D{=}4$ speculative-only decoding scores $45.17$ against SFT's $45.95$ with $1.7\times$ throughput, whereas $D{=}8$ hybrid decoding reaches $2.2\times$ at $43.25$. On Bonsai-4B, speculative-only decoding achieves a mean score of $64.13$, compared with $63.55$ for SFT, while retaining a $1.9\times$ throughput speedup.

{\looseness=-1
Increasing $D$ from $4$ to $8$ raises hybrid mean throughput from $722.12$ to $805.57$ tokens/s on BitNet and from $952.18$ to $1110.65$ on Bonsai-1.7B. Useful progress depends on the committed prefix rather than the proposed width alone. For Bonsai-1.7B hybrid decoding at $D{=}8$, BitMTP obtains $2.7\times$ and $3.0\times$ speedups on HumanEval+ and MBPP+, and a $1.7\times$ speedup on IFEval. These runs commit approximately $6.7$ and $7.1$ tokens per iteration on the code tasks, versus $4.6$ on IFEval. This pattern is consistent with amortizing block computation and state updates over more committed tokens, connecting workload-dependent acceleration to the execution state managed by BitMTP. With $D{=}8$ hybrid decoding, both Bonsai-1.7B and Bonsai-4B achieve more than twice the five-task mean throughput of their respective SFT baselines while maintaining comparable mean task quality. These results show that BitMTP retains substantial acceleration as the backbone scales from 1.7B to 4B parameters.
\par}

Tables~\ref{tab:portability_3090} and~\ref{tab:edge_orin} extend the evaluation to consumer and edge GPUs. On RTX~3090, $D{=}8$ hybrid decoding achieves five-task mean speedups of $1.9\times$ on BitNet and $2.0\times$ on Bonsai-1.7B. On Jetson Orin's MBPP+ workload, BitNet throughput increases from $19.35$ to $61.76$ tokens/s and Bonsai from $15.95$ to $52.05$, giving $3.2\times$ and $3.3\times$ speedups. All three policies accelerate MBPP+ over SFT AR on Orin, with hybrid decoding the fastest at both block sizes. Together, these results demonstrate effective multi-token acceleration across server, consumer, and edge GPUs, showing that the benefits of BitMTP extend beyond a single hardware class.

\begin{table}[t]
\centering
\caption{Five-task quality (Sc., \%) and decode throughput (Rate, tokens/s) of hybrid decoding on NVIDIA RTX~3090. Parentheses give speedups over SFT AR; Average is the unweighted task mean.}
\label{tab:portability_3090}
\scriptsize
\renewcommand{\submetric}[1]{\,{\fontsize{6.1pt}{7pt}\selectfont\speedup{(#1)}}}
\setlength{\tabcolsep}{1.5pt}
\renewcommand{\arraystretch}{1.10}
\resizebox{\linewidth}{!}{%
\begin{tabular}{@{}l *{5}{cc} cc@{}}
\toprule
\multirowcell{2}{Configuration}
& \multicolumn{2}{c}{GSM8K} & \multicolumn{2}{c}{MATH-500} & \multicolumn{2}{c}{IFEval}
& \multicolumn{2}{c}{HumanEval+} & \multicolumn{2}{c}{MBPP+} & \multicolumn{2}{c}{Average}\\
\cmidrule(lr){2-3}\cmidrule(lr){4-5}\cmidrule(lr){6-7}\cmidrule(lr){8-9}\cmidrule(lr){10-11}\cmidrule(lr){12-13}
& Sc. & Rate & Sc. & Rate & Sc. & Rate & Sc. & Rate & Sc. & Rate & Sc. & Rate\\
\midrule
\mbox{BitNet~b1.58-2B-SFT} & 58.38 & 275.54 & 35.00 & 281.68 & 56.35 & 277.36 & 39.02 & 284.19 & 40.74 & 282.94 & 45.90 & 280.34\\
\mbox{BitMTP ($D{=}4$)} & 62.77 & 510.62\submetric{1.9$\times$} & 35.00 & 492.56\submetric{1.7$\times$} & 54.08 & 392.57\submetric{1.4$\times$} & 35.37 & 572.04\submetric{2.0$\times$} & 33.60 & 553.23\submetric{2.0$\times$} & 44.16 & 504.20\submetric{1.8$\times$}\\
\mbox{BitMTP ($D{=}8$)} & 59.89 & 524.16\submetric{1.9$\times$} & 33.00 & 515.08\submetric{1.8$\times$} & 55.76 & 364.84\submetric{1.3$\times$} & 35.37 & 661.16\submetric{2.3$\times$} & 33.33 & 626.82\submetric{2.2$\times$} & 43.47 & 538.41\submetric{1.9$\times$}\\
\midrule
\mbox{Bonsai-1.7B-SFT} & 61.18 & 244.51 & 42.00 & 245.31 & 68.94 & 244.81 & 35.37 & 245.08 & 36.51 & 244.64 & 48.80 & 244.87\\
\mbox{BitMTP ($D{=}4$)} & 57.54 & 430.60\submetric{1.8$\times$} & 42.60 & 472.35\submetric{1.9$\times$} & 67.39 & 376.04\submetric{1.5$\times$} & 34.15 & 516.57\submetric{2.1$\times$} & 37.57 & 474.59\submetric{1.9$\times$} & 47.85 & 454.03\submetric{1.9$\times$}\\
\mbox{BitMTP ($D{=}8$)} & 58.00 & 416.82\submetric{1.7$\times$} & 42.40 & 507.39\submetric{2.1$\times$} & 66.91 & 346.49\submetric{1.4$\times$} & 31.71 & 580.83\submetric{2.4$\times$} & 37.83 & 639.76\submetric{2.6$\times$} & 47.37 & 498.26\submetric{2.0$\times$}\\
\bottomrule
\end{tabular}}

\end{table}

\begin{table}[t]
\centering
\caption{MBPP+ decode throughput (tokens/s) on Jetson Orin. Parentheses give speedups over SFT AR; decoding-policy abbreviations follow Table~\ref{tab:gen_bench_main}.}
\label{tab:edge_orin}
\scriptsize
\setlength{\tabcolsep}{4pt}
\renewcommand{\arraystretch}{1.08}
\begin{tabular*}{\linewidth}{@{\extracolsep{\fill}}lccccccc@{}}
\toprule
\multirowcell{2}{Backbone} & \multirowcell{2}{SFT AR} & \multicolumn{3}{c}{BitMTP ($D{=}4$)} & \multicolumn{3}{c}{BitMTP ($D{=}8$)}\\
\cmidrule(lr){3-5}\cmidrule(lr){6-8}
& & Conf. & Spec. & Hybr. & Conf. & Spec. & Hybr.\\
\midrule
BitNet~b1.58-2B
& 19.35\submetric{1.0$\times$}
& 35.38\submetric{1.8$\times$}
& 34.38\submetric{1.8$\times$}
& 40.92\submetric{2.1$\times$}
& 44.74\submetric{2.3$\times$}
& 54.55\submetric{2.8$\times$}
& 61.76\submetric{3.2$\times$}\\
Bonsai-1.7B
& 15.95\submetric{1.0$\times$}
& 30.80\submetric{1.9$\times$}
& 28.83\submetric{1.8$\times$}
& 36.20\submetric{2.3$\times$}
& 40.42\submetric{2.5$\times$}
& 42.47\submetric{2.7$\times$}
& 52.05\submetric{3.3$\times$}\\
\bottomrule
\end{tabular*}
\end{table}

\begin{table}[t]
\centering
\caption{Eleven-task model capability scores (\%). Avg. denotes the unweighted mean; TQA is TruthfulQA-mc2. W/A denotes weight/activation precision. TriviaQA scores generated answers; the remaining tasks use likelihood scoring.}
\label{tab:likelihood_bench}
\footnotesize
\setlength{\tabcolsep}{2.2pt}
\renewcommand{\arraystretch}{1.10}
\resizebox{\linewidth}{!}{%
\begin{tabular}{lccccccccccccc}
\toprule
Model & W/A & MMLU & Wino & Hella & ARC-C & ARC-E & BoolQ & OBQA & PIQA & Trivia & CSQA & TQA & Avg.\\
\midrule
MiniCPM-2B & BF16/BF16 & 52.4 & 65.5 & 71.4 & 43.8 & 68.9 & 77.6 & 42.6 & 76.1 & 34.2 & 73.1 & 43.1 & 59.0\\
Qwen2.5-1.5B & BF16/BF16 & 60.2 & 63.4 & 68.3 & 46.6 & 75.9 & 78.1 & 40.4 & 75.9 & 39.0 & 76.5 & 46.7 & 61.0\\
Llama-3.2-3B & BF16/BF16 & 60.7 & 68.6 & 71.6 & 46.1 & 71.0 & 78.8 & 39.2 & 76.9 & 58.6 & 72.5 & 51.5 & 63.2\\
\midrule
Qwen2.5-1.5B-AWQ & INT4/FP16 & 57.5 & 63.5 & 66.4 & 44.3 & 72.5 & 74.4 & 40.4 & 75.6 & 33.9 & 74.0 & 45.4 & 58.9\\
Llama3-8B-1.58$^{\dagger}$ & 1.58/BF16 & 35.2 & 59.2 & 68.7 & 43.4 & 69.8 & 68.6 & 39.4 & 75.9 & 20.0 & 30.0 & 39.1 & 49.9\\
Falcon3-7B-1.58bit$^{\dagger}$ & 1.58/BF16 & 43.3 & 60.1 & 59.5 & 37.6 & 64.8 & 72.3 & 38.6 & 72.8 & 0.0 & 67.3 & 43.2 & 50.9\\
BitCPM4-1B$^{\dagger}$ & 1.58/BF16 & 55.2 & 55.8 & 52.6 & 38.9 & 63.0 & 70.4 & 36.6 & 69.1 & 15.4 & 55.0 & 42.5 & 50.4\\
\midrule
\multicolumn{14}{c}{\textit{BitNet~b1.58-2B family}}\\
\midrule
BitNet b1.58-2B & 1.58/INT8 & 53.0 & 72.1 & 67.6 & 48.1 & 73.7 & 79.2 & 40.0 & 76.6 & 33.7 & 71.8 & 46.4 & 60.2\\
BitNet b1.58-2B-SFT & 1.58/INT8 & 53.2 & 69.6 & 67.8 & 50.6 & 76.1 & 79.5 & 42.4 & 78.1 & 33.6 & 69.6 & 39.8 & 60.0\\
\mbox{BitMTP ($D{=}4$)} & 1.58/INT8 & 52.9 & 70.2 & 67.2 & 48.8 & 75.1 & 79.1 & 41.8 & 78.0 & 33.0 & 70.3 & 41.7 & 59.8\\
\mbox{BitMTP ($D{=}8$)} & 1.58/INT8 & 52.0 & 70.0 & 66.0 & 48.6 & 76.5 & 78.7 & 39.8 & 77.2 & 29.8 & 69.4 & 42.8 & 59.1\\
\midrule
\multicolumn{14}{c}{\textit{Bonsai-1.7B family}}\\
\midrule
Bonsai-1.7B & 1.58/INT8 & 49.2 & 59.1 & 52.4 & 38.5 & 65.2 & 72.4 & 33.8 & 70.1 & 19.8 & 63.4 & 46.9 & 51.9\\
Bonsai-1.7B-SFT & 1.58/INT8 & 46.1 & 58.0 & 53.8 & 37.8 & 64.1 & 71.7 & 33.6 & 69.4 & 19.2 & 60.4 & 43.4 & 50.7\\
\mbox{BitMTP ($D{=}4$)} & 1.58/INT8 & 43.0 & 55.7 & 52.3 & 35.6 & 64.5 & 70.4 & 32.2 & 70.6 & 17.7 & 56.7 & 42.7 & 49.2\\
\mbox{BitMTP ($D{=}8$)} & 1.58/INT8 & 44.0 & 56.7 & 52.3 & 36.2 & 63.8 & 74.4 & 33.8 & 70.2 & 17.5 & 59.9 & 41.8 & 50.0\\
\bottomrule
\end{tabular}}

\par\smallskip\noindent\begin{minipage}{\linewidth}\footnotesize
$^{\dagger}$BF16 reference execution.
\end{minipage}
\end{table}

\begin{table}[t]
\centering
\caption{BitNet hybrid-decoding configurations at $D{\in}\{4,8\}$. Rates are decode tokens/s; Avg. is the unweighted task mean. Parentheses give speedups over SFT AR; HE+ denotes HumanEval+.}
\label{tab:ablation_main}
\footnotesize
\setlength{\tabcolsep}{2.0pt}
\renewcommand{\arraystretch}{1.10}
\resizebox{\linewidth}{!}{%
\begin{tabular}{l cccccc cccccc}
\toprule
\multirowcell{2}{Runtime\\configuration}
& \multicolumn{6}{c}{$D{=}4$ (tokens/s)}
& \multicolumn{6}{c}{$D{=}8$ (tokens/s)}\\
\cmidrule(lr){2-7}\cmidrule(lr){8-13}
& GSM8K & MATH-500 & IFEval & HE+ & MBPP+ & Avg.
& GSM8K & MATH-500 & IFEval & HE+ & MBPP+ & Avg.\\
\midrule
SFT AR & 365.0 & 365.0 & 368.7 & 393.7 & 374.8 & 373.4\submetric{1.0$\times$} & 365.0 & 365.0 & 368.7 & 393.7 & 374.8 & 373.4\submetric{1.0$\times$}\\
\midrule
Generic & 158.0 & 133.0 & 103.8 & 157.5 & 162.9 & 143.0\submetric{0.4$\times$} & 268.9 & 215.9 & 144.2 & 290.5 & 301.3 & 244.2\submetric{0.7$\times$}\\
$+\,$M1 & 724.2 & 658.2 & 515.1 & 739.2 & 758.0 & 678.9\submetric{1.8$\times$} & 759.1 & 663.7 & 429.6 & 793.7 & 840.6 & 697.4\submetric{1.9$\times$}\\
$+\,$M1$+$M2 & 735.2 & 664.8 & 529.1 & 746.0 & 772.5 & 689.5\submetric{1.8$\times$} & 784.5 & 684.3 & 452.4 & 825.3 & 879.6 & 725.2\submetric{1.9$\times$}\\
$+\,$M1$+$M2$+$M3 (Full) & 761.4 & 707.1 & 580.4 & 768.7 & 793.0 & 722.1\submetric{1.9$\times$} & 819.4 & 777.0 & 543.6 & 923.6 & 964.2 & 805.6\submetric{2.2$\times$}\\
\bottomrule
\end{tabular}}
\vspace{+1mm}
\end{table}

\subsection{Model Capability Retention}
\label{sec:exp_likelihood}
Table~\ref{tab:likelihood_bench} compares BitMTP with the SFT counterpart of each backbone to assess retention of knowledge, reasoning, and commonsense capabilities beyond the generation benchmarks. Ten tasks assess the adapted backbone through causal likelihood scoring, without invoking multi-token \mbox{acceptance policies}; TriviaQA evaluates generated answers by exact match. Relative to SFT, \mbox{BitNet's} eleven-task means remain close at $59.8$ and $59.1$ for $D{=}4$ and $D{=}8$, compared with $60.0$. Bonsai-1.7B scores $49.2$ and $50.0$, compared with $50.7$ for SFT. The block-size ordering differs across backbones: $D{=}4$ has the higher eleven-task mean on BitNet, whereas $D{=}8$ scores higher on Bonsai-1.7B. On both backbones, BitMTP maintains eleven-task mean scores close to SFT, with differences of \mbox{$0.2$--$1.5$ percentage points.}

BitNet at $D{=}4$ scores higher than SFT on WinoGrande and TruthfulQA-mc2, while Bonsai-1.7B at $D{=}8$ scores $74.4$ versus $71.7$ on BoolQ but lower on MMLU and HellaSwag. The aggregate scores thus reflect a task-dependent pattern of retained strengths, gains, and reductions. Relative to compact-model baselines, BitNet's MTP means remain near MiniCPM-2B's $59.0$ and Qwen2.5-1.5B-AWQ's $58.9$, while BF16 Qwen2.5-1.5B and Llama-3.2-3B score higher at $61.0$ and $63.2$. Together with Section~\ref{sec:exp_main}, these results provide complementary evidence for BitMTP's generation acceleration and broad \mbox{capability retention relative to SFT.}

\subsection{Runtime Analysis}
\label{sec:exp_ablation}
\textbf{Decode efficiency.} Table~\ref{tab:ablation_main} compares different runtime configurations. The Generic configuration reaches a five-task mean throughput of $143.0$ tokens/s at $D{=}4$ and $244.2$ tokens/s at $D{=}8$. Both rates are below the SFT AR baseline of $373.42$ tokens/s. The Full configuration reaches $722.12$ tokens/s at $D{=}4$ and $805.57$ tokens/s at $D{=}8$. It achieves the highest throughput on all five tasks at both block sizes. At $D{=}8$, its task-level speedups range from $1.5\times$ on IFEval to $2.6\times$ on MBPP+. These gains span mathematical reasoning, \mbox{instruction following, and code generation.}

\textbf{Acceptance and runtime cost.} In the forced-rejection test reported in \textbf{Appendix~\ref{app:additional_analyses}}, each round performs block drafting and verification but commits at most one verifier-origin token, resulting in throughput below AR. In contrast, evaluated Bonsai-1.7B hybrid decoding at $D{=}8$ commits $4.6$--$7.1$ tokens per iteration across five tasks. Multi-token commitment amortizes block execution and verification. \textbf{Appendix~\ref{app:cross_device}} quantifies the memory and initialization costs of deploying the full runtime. The $0.5$--$1.6$~MiB shared-KV increment on H20 supports history reuse across replay graphs, while Orin's one-time startup overhead is amortized \mbox{in persistent serving.} Therefore, the proposed BitMTP translates multi-token commitment into decode speedups while sharing \mbox{causal history across replay states.}

\section{Conclusion and Limitation}
\label{sec:conclusion}
BitMTP enabled efficient multi-token generation on ultra-low-bit LLM backbones through joint NTP$+$MTP training, three decoding policies, and accepted-prefix-aligned runtime execution. The runtime combined width-specialized operators, shared-KV graph replay, and verifier scheduling. At $D{=}8$ on H20, hybrid decoding reached five-task mean speedups of $2.2\times$ on BitNet~b1.58-2B and $2.1$--$2.4\times$ on 1.7B--4B Bonsai models at comparable task quality. BitMTP reached up to $2.0\times$ in five-task mean throughput on RTX~3090 and $3.3\times$ on MBPP+ on Jetson Orin. These results demonstrated effective low-bit multi-token acceleration across ternary backbones, \mbox{model scales, and GPU types.}

There are some limitations of BitMTP. First, while BitMTP has been evaluated on 1.7B--4B backbones, its quality--throughput tradeoffs on larger models and longer contexts still need validation. Second, the current runtime relies on CUDA-specific low-bit kernels and CUDA Graphs and cannot run directly on non-CUDA accelerators. We will investigate these problems in further work.

{\small
\bibliography{references}
}

\clearpage
\appendix
\setcounter{topnumber}{3}
\setcounter{totalnumber}{4}
\renewcommand{\topfraction}{0.88}
\renewcommand{\textfraction}{0.10}
\renewcommand{\floatpagefraction}{0.75}
\setlength{\floatsep}{10pt plus 0pt minus 2pt}
\makeatletter
\setlength{\@fptop}{0pt}
\setlength{\@fpsep}{12pt}
\setlength{\@fpbot}{0pt plus 1fil}
\makeatother
\setcounter{equation}{0}
\renewcommand{\theequation}{\thesection.\arabic{equation}}
\renewcommand{\theHequation}{\thesection.\arabic{equation}}

\begin{center}
{\LARGE\bfseries Appendix}
\end{center}
\label{app:supp}

\noindent In this appendix, we provide the following materials:
\begin{itemize}[leftmargin=2.0em, topsep=2pt, itemsep=1pt, parsep=0pt]
\item[\textbf{A.}] Method and runtime details, including one hybrid decoding iteration, verifier and KV-state transitions, and replay-correctness requirements (referring to Section~\ref{sec:method} in the main paper).
\item[\textbf{B.}] Training data and cost (referring to Section~\ref{sec:exp_setup} in the main paper).
\item[\textbf{C.}] Forced-rejection analysis for the runtime configuration study (referring to Section~\ref{sec:exp_ablation} \mbox{in the main paper}).
\item[\textbf{D.}] GPU-resident model and runtime memory, graph-state overhead, and Jetson Orin cold-start measurements (referring to \mbox{Section~\ref{sec:exp_ablation} in the main paper}).
\end{itemize}

\section{Method and Runtime Details}
\label{app:method_details}

Algorithm~\ref{alg:bitmtp_iteration} \mbox{summarizes one hybrid decoding iteration.} The draft replay computes causal KV for the pending suffix while proposing a new block, and the runtime copies the pending-token KV rows into history. Hybrid decoding commits a sufficiently long confidence prefix; otherwise, an AR-causal verifier processes $[\text{anchor},d_1,\ldots,d_D]$ at width $D{+}1$ and determines the accepted prefix. Its additional token corrects the first rejected position or serves as a bonus after full acceptance. Confidence-only decoding uses the confidence branch, while speculative-only decoding \mbox{always uses the verifier branch.}

\begin{algorithm}[H]
\caption{Hybrid decoding state update.}
\label{alg:bitmtp_iteration}
\small
\begin{algorithmic}[1]
\Require state $q=(p,\ell,u)$ with $u\ne[]$, block size $D$, and controls $(\tau,m)$
\State set $n\leftarrow|u|$ and select $s_{\mathrm{d}}=(M,\sigma,r_{\mathrm{d}})$ with $P=M-D\ge n$
\State form $x_{\mathrm{d}}=[u,\mathrm{PAD}^{P-n},\operatorname{last}(u),m_{\mathrm{mask}}^{D-1}]$
\State replay the draft graph on $x_{\mathrm{d}}$ and obtain $d_{1:D}$ from the final $D$ output rows
\State materialize only the $n$ pending KV rows and set $\bar\ell\leftarrow\ell+n$
\State compute $h$ on the GPU using Eq.~\ref{eq:confidence_accept}
\State select $s_{\mathrm{v}}=(D{+}1,\sigma,r_{\mathrm{v}})$ and prepare $[\operatorname{last}(u),d_{1:D}]$
\State prepare verifier positions and the AR-causal mask on the GPU stream
\State obtain $h$ on the host
\If{$h\ge m$}
  \State set $\hat a\leftarrow h$
  \State append $d_{1:\hat a}$ to $p$
  \State set $u'\leftarrow d_{1:\hat a}$ and $\ell'\leftarrow\bar\ell$
\Else
  \State replay the verifier graph for $s_{\mathrm{v}}$ and obtain $\tilde d_{1:D+1}$
  \State set $\hat a$ to the verifier-accepted prefix length
  \State set $b\leftarrow\tilde d_{\hat a+1}$
  \State promote the $\hat a$ accepted draft KV rows, excluding the anchor
  \State append $d_{1:\hat a}\!\cdot\!b$ to $p$
  \State set $u'\leftarrow[b]$ and $\ell'\leftarrow\bar\ell+\hat a$
\EndIf
\State return $q'=(p,\ell',u')$
\end{algorithmic}
\end{algorithm}

After a confidence transition, the committed drafts remain pending in the updated state until a later causal replay computes their KV from the emitted token ids. A verifier transition instead promotes the accepted causal rows and leaves only the verifier-origin token pending. Both branches preserve Eq.~\ref{eq:prefix_invariant}. In the verifier branch, full acceptance appends the $(D{+}1)$-st verifier token as a bonus, whereas rejection at the first draft position emits only \mbox{the verifier correction.}

Superscripts on $\mathrm{PAD}$ and $m_{\mathrm{mask}}$ denote repeated input slots. The draft anchor repeats the final pending token and its position; block queries hide its original pending copy. Verifier positions start at $\bar\ell-1$, and its mask hides the anchor's history copy. Padding, mask, and rejected draft rows do not enter committed causal history. Each reachable pending length $n$ has a captured draft width $M\ge n+D$; verification uses width $D{+}1$. Graphs of different widths and roles share $B_\sigma$, with history migration only when capacity grows. These choices preserve \mbox{the prefix-length relation in Eq.~\ref{eq:prefix_invariant}.}

\begin{table}[t]
\centering
\caption{Composition of the processed training mixture used in the training-cost study.}
\label{tab:supp-training-data-mixture}
\small
\setlength{\tabcolsep}{5pt}
\renewcommand{\arraystretch}{1.08}
\begin{tabular*}{\linewidth}{@{\extracolsep{\fill}}lrl@{}}
\toprule
\textbf{Source} & \textbf{Examples} & \textbf{Domain} \\
\midrule
ScaleQuest-Math & 1{,}003{,}467 & Math \\
SmolTalk & 1{,}038{,}618 & General \\
OPC-SFT-Stage2 & 435{,}930 & Code \\
Tulu-3-SFT-Mixture & 923{,}278 & General \\
SciRIFF & 70{,}520 & Science \\
Table-GPT & 13{,}214 & Table \\
OpenCodeInstruct & 17{,}494 & Code \\
Reasoning-SFT-IF & 41{,}413 & \mbox{Instruction following} \\
\midrule
\textbf{Total} & \textbf{3{,}543{,}934} & N/A \\
\bottomrule
\end{tabular*}
\end{table}

\begin{table}[t]
\centering
\caption{One-epoch training cost on Bonsai-1.7B using eight NVIDIA H20 GPUs. All configurations use the same corpus, global batch size, and optimizer-step count. Relative cost uses the one-epoch SFT run as its baseline.}
\label{tab:training_cost}
\small
\setlength{\tabcolsep}{5pt}
\renewcommand{\arraystretch}{1.12}
\begin{tabular*}{\linewidth}{@{\extracolsep{\fill}}lrrr@{}}
\toprule
Configuration & Wall time (h) & GPU-hours & Relative cost\\
\midrule
SFT & 71.29 & 570.3 & $1.00\times$\\
BitMTP ($D{=}4$) & 103.10 & 824.8 & $1.45\times$\\
BitMTP ($D{=}8$) & 112.88 & 903.0 & $1.58\times$\\
\bottomrule
\end{tabular*}
\end{table}

\section{Training Data and Cost}
\label{app:training_data}

Table~\ref{tab:supp-training-data-mixture} lists the eight training sources, covering mathematics, general dialogue, code, science, tables, and instruction following. For each backbone, SFT and joint NTP$+$MTP training use the same processed mixture drawn from these sources, comprising \mbox{approximately 3.5 million examples.}

Table~\ref{tab:training_cost} reports one-epoch training cost on Bonsai-1.7B using eight H20 GPUs. With the corpus and optimizer-step count fixed, joint NTP$+$MTP training uses $1.45\times$/$1.58\times$ the SFT GPU-hours at $D{=}4$/$8$, quantifying the additional cost of \mbox{joint AR and block-prediction supervision.}

\section{Additional Experimental Analyses}
\label{app:additional_analyses}

\begin{table}[t]
\centering
\caption{Forced-rejection stress test on H20. Decode gives five-task mean throughput (tokens/s); dark-red values in parentheses show speedups over SFT AR. Accepted and Com. report per-iteration means of accepted draft tokens and committed output tokens.}
\label{tab:forced_rejection}
\scriptsize
\setlength{\tabcolsep}{5.0pt}
\renewcommand{\arraystretch}{1.12}
\begin{tabular*}{\linewidth}{@{\extracolsep{\fill}}l c ccc ccc@{}}
\toprule
\multirowcell{2}{\textbf{Backbone}} & \multirowcell{2}{\textbf{SFT AR}\\\textbf{Decode}}
& \multicolumn{3}{c}{\textbf{Forced rejection, $D{=}4$}}
& \multicolumn{3}{c}{\textbf{Forced rejection, $D{=}8$}}\\
\cmidrule(lr){3-5}\cmidrule(lr){6-8}
& & Decode & Accepted & Com. & Decode & Accepted & Com.\\
\midrule
BitNet~b1.58-2B & 373.42\submetric{1.0$\times$} & 142.91\submetric{0.4$\times$} & 0 & 0.9975 & 103.40\submetric{0.3$\times$} & 0 & 0.9974\\
Bonsai-1.7B & 469.26\submetric{1.0$\times$} & 188.57\submetric{0.4$\times$} & 0 & 0.9978 & 160.28\submetric{0.3$\times$} & 0 & 0.9978\\
\bottomrule
\end{tabular*}
\end{table}

\textbf{Forced-rejection stress test.}\label{app:forced_rejection}
The stress test in Table~\ref{tab:forced_rejection} forces zero acceptance in each draft--verify round, paying for $D$ drafts and width-$(D{+}1)$ verification while committing at most one verifier-origin token. The result is below AR, with lower throughput at $D{=}8$ than at $D{=}4$ on both backbones. Individual zero-acceptance rounds can occur normally, but this all-reject stress test is an artificial worst case. Evaluated Bonsai-1.7B hybrid workloads at $D{=}8$ commit $4.6$--$7.1$ tokens per iteration, amortizing drafting and verification to support \mbox{the measured low-bit generation speedups.}

\section{Runtime Memory and Startup Cost}
\label{app:cross_device}

\begin{table}[t]
\centering
\caption{H20 model and runtime memory at a 4{,}096-token segment, in MiB. Model total includes Emb./head. Static contains per-graph token, position, and mask buffers. Other allocated is CUDA-allocated memory after graph capture minus Model total, Shared KV, and Static. Peak is the median of three runs' maximum device-memory readings across the recorded stages.}
\label{tab:h20_memory_breakdown}
\footnotesize
\setlength{\tabcolsep}{2.4pt}
\renewcommand{\arraystretch}{1.12}
\resizebox{\linewidth}{!}{%
\begin{tabular}{lrrrrrrrrr}
\toprule
Configuration & Graphs & Emb./head & BF16 prefill & Packed decode & Model total & Shared KV & Static & \makecell{Other\\allocated} & Peak\\
\midrule
\multicolumn{10}{c}{\textit{BitNet~b1.58-2B}}\\
\midrule
SFT AR & 1 & 2505.0 & 6563.3 & 881.6 & 9950.0 & 300.1 & 0.5 & 132.4 & 11028\\
BitMTP ($D{=}4$) & 7 & 2818.1 & 6563.3 & 881.6 & 10263.1 & 300.6 & 13.7 & 326.8 & 11714\\
BitMTP ($D{=}8$) & 7 & 2818.1 & 6563.3 & 881.6 & 10263.1 & 301.2 & 26.9 & 327.6 & 11742\\
\midrule
\multicolumn{10}{c}{\textit{Bonsai-1.7B}}\\
\midrule
SFT AR & 1 & 671.3 & 0.0 & 651.2 & 1322.5 & 448.1 & 0.6 & 0.0 & 5118\\
BitMTP ($D{=}4$) & 7 & 967.5 & 0.0 & 651.2 & 1618.8 & 448.9 & 16.1 & 165.1 & 5656\\
BitMTP ($D{=}8$) & 7 & 967.5 & 0.0 & 651.2 & 1618.8 & 449.8 & 31.5 & 165.1 & 5656\\
\bottomrule
\end{tabular}}

\medskip
\centering
\normalsize
\caption{Jetson Orin memory and cold-start cost at a 4{,}096-token segment. Memory is in MiB. Capture is the allocator increase from KV allocation to ready-to-serve. Peak reserved is the first-request CUDA allocator peak; cold start uses an empty compiler cache.}
\label{tab:orin_cold_start}
\small
\setlength{\tabcolsep}{4pt}
\renewcommand{\arraystretch}{1.12}
\begin{tabular*}{\linewidth}{@{\extracolsep{\fill}}lrrrrr@{}}
\toprule
Configuration & Graphs & Shared KV & Capture & Peak reserved & Cold start (s)\\
\midrule
BitNet~b1.58-2B-SFT AR & 1 & 301.2 & 16.7 & 11312 & 14.34\\
BitMTP--BitNet ($D{=}4$) & 7 & 300.7 & 79.2 & 11716 & 17.30\\
BitMTP--BitNet ($D{=}8$) & 7 & 301.2 & 93.4 & 11730 & 17.92\\
\midrule
Bonsai-1.7B-SFT AR & 1 & 448.1 & 16.8 & 5486 & 13.41\\
BitMTP--Bonsai ($D{=}4$) & 7 & 449.0 & 81.6 & 5876 & 16.52\\
BitMTP--Bonsai ($D{=}8$) & 7 & 449.9 & 97.9 & 5960 & 16.53\\
\bottomrule
\end{tabular*}
\end{table}

Tables~\ref{tab:h20_memory_breakdown} and~\ref{tab:orin_cold_start} quantify memory and startup costs at a 4{,}096-token segment. SFT AR captures one graph and BitMTP captures seven graphs over shared causal KV. Model totals include all resident weight representations and \mbox{the embedding and output-projection storage.}

H20 shared KV increases by only $0.5$--$1.6$~MiB from one to seven graphs, reusing causal history rather than duplicating full caches. Sampled peak device memory is 686--714~MiB higher on BitNet and 538~MiB higher on Bonsai-1.7B \mbox{than on SFT AR.}

Orin's first-request peak reserved memory increases by 404--418~MiB on BitNet and 390--474~MiB on Bonsai-1.7B, with cold-start increments of 2.96--3.58~s and 3.11--3.12~s, respectively. This one-time per-process cost is amortized over requests, supporting persistent multi-token generation through \mbox{shared history and replayable execution states.}

\end{document}